\documentclass[conference]{IEEEtran}
\usepackage{cite}
\usepackage{amsmath,amssymb,amsfonts}
\usepackage{algorithmic}
\usepackage{graphicx}
\usepackage{multicol} % If you're using multicol for columns
\usepackage{textcomp}
\usepackage{xcolor}
\def\BibTeX{{\rm B\kern-.05em{\sc i\kern-.025em b}\kern-.08em
    T\kern-.1667em\lower.7ex\hbox{E}\kern-.125emX}}
\begin{document}

\title{Beyond Sub-6 GHz: Leveraging mmWave Wi-Fi for Gait-Based Person Identification\\
}

\author{
    \IEEEauthorblockN{Nabeel Nisar Bhat\IEEEauthorrefmark{1}, Maksim Karnaukh\IEEEauthorrefmark{2}, Jakob Struye\IEEEauthorrefmark{1}, Rafael Berkvens\IEEEauthorrefmark{1}, Jeroen Famaey\IEEEauthorrefmark{1}}
    \IEEEauthorblockA{
        \IEEEauthorrefmark{1}\textit{IDLab}, \textit{University of Antwerp -- imec}, Antwerp, Belgium\\
        Email: \{firstname.lastname\}@uantwerpen.be}
    \IEEEauthorblockA{
        \IEEEauthorrefmark{2}\textit{Faculty of Science}, \textit{University of Antwerp}, Antwerp, Belgium\\
        Email: maksim.karnaukh@student.uantwerpen.be}}
\maketitle

\begin{abstract}

Person identification plays a vital role in enabling intelligent, personalized, and secure human-computer interaction. Recent research has demonstrated the feasibility of leveraging Wi-Fi signals for passive person identification using a person's unique gait pattern. Although most existing work focuses on sub-6 GHz frequencies, the emergence of mmWave offers new opportunities through its finer spatial resolution, though its comparative advantages for person identification remain unexplored. 
This work presents the first comparative study between sub-6 GHz and mmWave Wi-Fi signals for person identification with commercial-off-the-shelf (COTS) Wi-Fi, using a novel dataset of synchronized measurements from the two frequency bands in an indoor environment.
To ensure a fair comparison, we apply identical training pipelines and model configurations across both frequency bands. 
Leveraging end-to-end deep learning, we show that even at low sampling rates (10 Hz), mmWave Wi-Fi signals can achieve high identification accuracy (91.2\% on 20 individuals) when combined with effective background subtraction.

\end{abstract}


\section{Introduction}

In recent years, Wi-Fi-based human identification has gained traction as a promising, non-intrusive approach, driven by the widespread availability of Wi-Fi devices and the need for passive authentication in smart environments. Traditional biometrics like fingerprints \cite{nguyen2024deep}, iris scans \cite{nguyen2024deep}, and facial recognition offer high accuracy but often require dedicated hardware, active participation, or raise privacy concerns. Alternatives such as wearable-based \cite{mekruksavanich2021biometric} and vision-based methods \cite{onyema2021enhancement} also face challenges, including lighting conditions, line-of-sight constraints, and similar privacy issues.

To overcome these limitations, researchers have leveraged the ubiquitous presence of Wi-Fi in indoor environments. As people move, their bodies uniquely alter RF signals through reflection, scattering, and absorption~\cite{rf_for_sensing2, wiwho}. These changes are captured via fine-grained Channel State Information (CSI), which characterizes signal propagation across subcarriers and antenna paths. Human gait has emerged as a particularly effective biometric as it involves full-body motion, is hard to spoof, and requires no user interaction or additional devices~\cite{harris2022survey}.

Building on these insights, systems like WiFi-ID \cite{wifi-id}, WiWho \cite{wiwho}, and WifiU \cite{wifiu} showed early promise by analyzing gait cycles and RF perturbations, but often relied on hand-crafted features and shallow classifiers, limiting scalability and robustness. Recent methods such as NeuralWave \cite{neuralwave}, HumanFi \cite{humanfi}, and Gate-ID \cite{gate-id} adopt deep learning to automatically extract discriminative features from raw CSI. NeuralWave uses a CNN-based architecture, HumanFi employs Long Short-Term Memory network (LSTM) for temporal modeling, and Gate-ID incorporates attention to handle directional gait and antenna placement, improving accuracy.

All of this research has focused on sub-6 GHz Wi-Fi, however, there is a growing interest in exploring the potential of mmWave Wi-Fi for human sensing and identification tasks. mmWave offers higher spatial resolution and finer signal granularity due to its shorter wavelength, which may enable a more precise capture of motion-induced variations in wireless signals. Most of the work in this domain currently relies on mmWave radar. For instance, Gu et al. proposed mmSense \cite{mmsense1}, a device-free system that uses 60 GHz mmWave radar to detect and identify multiple people based on environmental fingerprints and human body reflections, achieving person identification using LSTM-based models. Similarly, MGait \cite{mgait} employs a 77 GHz Frequency Modulated Continuous Wave (FMCW) radar to extract gait features from micro-Doppler signatures, enabling identification and intruder detection in multi-person environments with up to 88.6\% accuracy for 5 subjects.

% Other radar-based systems have combined multimodal data to improve recognition performance. An interesting example is the Cross Vision-RF Gait Re-identification framework \cite{crossvisionrf}, which fuses mmWave radar with RGB-D camera inputs to overcome modality gaps and achieve 92.5\% top-1 accuracy on a dataset of 56 participants.
However, radar-based mmWave systems frequently encounter obstacles such as high costs, complex hardware requirements, and limited availability, restricting their practical use in person identification. In contrast, some studies have explored the potential of using commercial off-the-shelf (COTS) Wi-Fi for gesture recognition \cite{nabeel1}. However, to the best of our knowledge, person identification using  mmWave COTS Wi-Fi remains significantly underexplored.

% Another study \cite{radarfusion} introduced a multimodal fusion algorithm that integrates mmWave radar-derived features such as phase, respiration, and heartbeat signals with an attention-based deep network, achieving 94.26\% accuracy on a dataset of 20 participants.

In this paper, our aim is to address this by conducting a comparative study of person identification performance using both traditional sub-6 GHz and mmWave Wi-Fi signals. Using end-to-end deep learning techniques, without the need for manual feature extraction, we analyze CSI data from both frequency bands to evaluate how well each system can distinguish people based on their gait. To enable this comparison, we collect a unique synchronized multi-modal CSI dataset using COTS hardware, with 5 GHz and 60 GHz measurements recorded simultaneously from 20 participants in an indoor setting. By systematically comparing the two, we investigate whether the higher resolution of mmWave offers measurable benefits for person identification, and we note the potential limitations and challenges.
% \subsection{Contributions}

The contributions of this work can be summarized as follows:

\begin{itemize}
    \item \textbf{Multi-modal CSI Dataset:} We address the lack of mmWave COTS datasets for person identification by collecting a comprehensive, synchronized multi-modal CSI dataset\footnote{https://ieee-dataport.org/documents/mmwavexr-multi-modal-and-distributed-mmwave-isac-datasets-human-sensing
} involving 20 individuals in an indoor environment. The dataset captures simultaneous 5 GHz and 60 GHz (mmWave) channel measurements, enabling the first direct comparison between sub-6 GHz and mmWave bands for human identification. This contribution fills a critical gap in the literature.
    \item \textbf{Cross-band Performance Comparison:} We present the first experimental comparison of mmWave and sub-6 GHz Wi-Fi CSI for person identification. Using several deep learning architectures in an end-to-end manner, we evaluate their accuracy under identical training conditions to better understand the trade-offs between the two technologies.
    \item \textbf{Benchmarking with Deep Models:} We implement and benchmark multiple temporal learning architectures including LSTMs, residual CNNs, and Temporal Convolutional Networks (TCNs). 
\end{itemize}

\section{Methodology} \label{sec:methodology}
This section discusses the dataset and the deep learning architectures employed for person identification.
\subsection{Dataset}
\begin{figure*}[!t]
\centering
\includegraphics[width=\textwidth,trim = 0cm 0cm 0cm 0cm, clip]{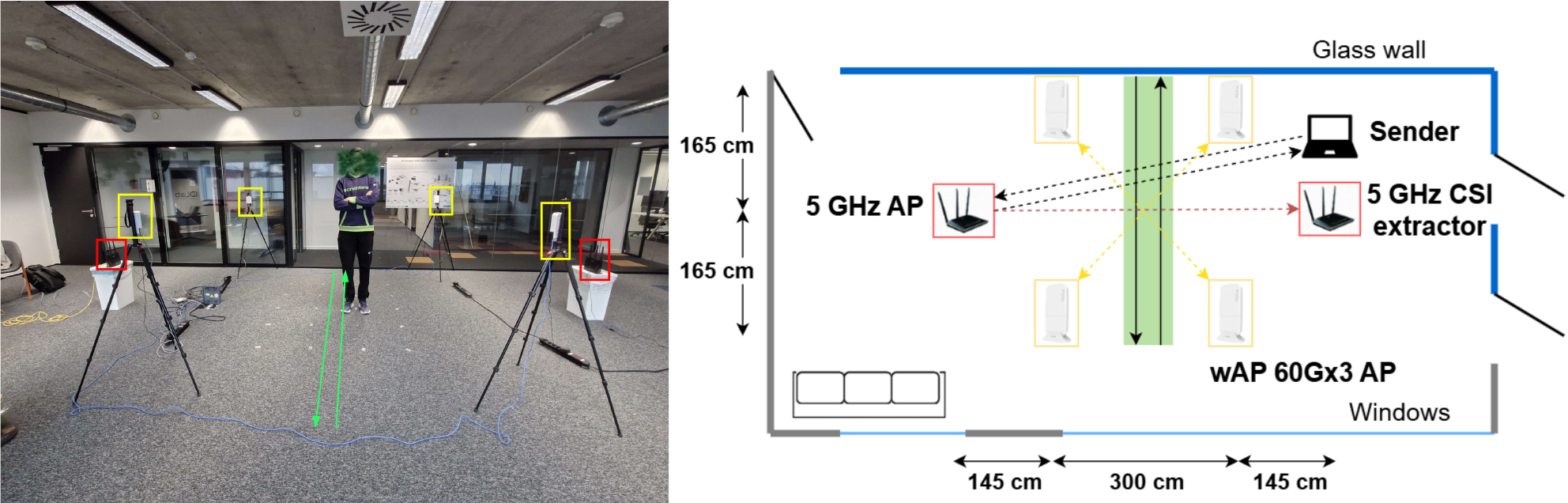}
\caption{Environment and experimental settings.}
\label{fig:settings}
\end{figure*}

We collect a novel multi-modal dataset capturing CSI from both 5 GHz and 60 GHz Wi-Fi devices, as illustrated in Figure \ref{fig:settings}. The data is recorded over three days in an indoor environment with 20 participants. Each participant walks along a linear path  (green area in  Figure \ref{fig:settings}) for two minutes per session, and background (no-person) recordings are taken on the final day. For sub-6 GHz, we use two ASUS RT-AC86U routers—one as an access point (AP) and the other as a passive CSI monitor. An Intel laptop transmits ICMP echo requests to the AP, while the monitor captures CSI responses in the 5 GHz band \cite{nabeel1}. To match the mmWave sampling rate, we downsample the 5 GHz CSI to 10 Hz (from 200 Hz). For 60 GHz, we deploy two independent MikroTik wAP 60Gx3 transmitter-receiver pairs arranged in a cross-shaped layout. CSI is captured at approximately 10 Hz from 30 antenna elements using an open-source tool \cite{blanco2022augmenting}. This setup enables synchronized person identification experiments across both frequency bands. The 60 GHz CSI has a shape of $26000 \times 30$ per device pair, while the 5 GHz CSI has a shape of $520000 \times 52$, where 52 represents the subcarriers. The data is segmented into 5-second chunks for input to downstream models.
% In the first setup, we focused solely on mmWave sensing. Using MikroTik wAP 60Gx3 devices and an open-source CSI extraction toolchain, we recorded amplitude-only CSI from 30 antenna elements at 22 frames per second. Seven participants walked back and forth between the devices for four minutes per round, completing two rounds each—resulting in eight minutes of data per individual.

\subsection{Data Preprocessing and Augmentation}

We adopt an end-to-end learning approach with minimal manual feature engineering. Raw CSI signals are fed directly into deep neural networks after a single preprocessing step: background subtraction~\cite{pegoraro2023rapid}, where a static CSI baseline (captured without any person) is subtracted to emphasize motion-induced variations and suppress environmental noise.

To improve generalization and reduce overfitting, we apply two data augmentation strategies during training: \textit{Gaussian smoothing} and \textit{Mixup}. Gaussian smoothing is applied probabilistically by convolving a 1D Gaussian filter along the temporal axis, reducing high-frequency noise and encouraging focus on stable motion patterns. The filter’s kernel size and standard deviation are tunable. Mixup~\cite{mixup_paper} creates synthetic samples by convexly combining input pairs and their labels, promoting linear behavior and robustness. When used together, smoothing is applied first to retain temporal structure before Mixup blends the inputs.

\subsection{Model Architectures}

We explore several deep learning models across recurrent and convolutional architectures. Inspired by prior sub-6 GHz CSI methods, we adapt and extend these models for the mmWave domain. All models are trained from scratch using supervised learning with amplitude-only CSI sequences as input, enabling end-to-end training without handcrafted features. We use a 70\%-15\%-15\% split for training, validation, and testing.

\subsubsection{LSTM and Variants}

We implement a LSTM model following the architecture proposed in the HumanFi paper~\cite{humanfi}, where a unidirectional LSTM processes the input signal and the hidden state of the final timestep is passed to a dense classifier. We extend this baseline in several ways:
\begin{itemize}
    \item \textbf{Bidirectional LSTM (BiLSTM)} to capture both past and future context.
    \item \textbf{CNN-BiLSTM with Attention}, which augments the BiLSTM with a CNN layer in front and attention mechanisms to selectively weight the temporal features. Variants include temporal attention and dual attention. The latter is based on the Zhu et al.~\cite{cnnbilstmatt} implementation.
\end{itemize}

\subsubsection{Residual CNN}

To exploit hierarchical temporal features in the CSI sequences, we look at the residual CNN (ResNet) inspired by the architecture used in work ~\cite{8740950}, originally proposed for joint activity recognition and indoor localization (JARIL). We explore several variants of this ResNet-style architecture:

% The base model consists of 4 layers, each composed of one residual block (i.e., a (1,1,1,1) configuration). Each block follows the classic two-layer bottleneck design with batch normalization and ReLU activation, and downsampling is applied at the beginning of each layer when needed to adjust the resolution or channel depth.
\begin{itemize}
    \item \textbf{OptResNet1D-JARIL:} A refined version of the original JARIL ResNet. We retain the (1,1,1,1) configuration, but we replace the aggressive initial downsampling with a more conservative stem (stride 1) to preserve fine-grained temporal details. The model includes four residual layers with progressively doubled channel widths, and the model ends with global average pooling and a fully connected output layer.
    \item \textbf{OptECAResNet1D-JARIL:} An extension of OptResNet1D-JARIL, enhanced with Efficient Channel Attention (ECA)~\cite{eca_paper} in each residual block. The ECA modules allow the network to adaptively reweight feature channels based on their temporal relevance, improving robustness to noise and focusing attention on discriminative gait features.
    \item \textbf{Custom ResNet:} A simplified variant with a consistent stride across all layers, reducing complexity and improving computational efficiency. We also evaluate an ECA-augmented version of this model.
\end{itemize}

\subsubsection{Temporal Convolutional Network (TCN)}

We also experiment with a TCN~\cite{tcn_paper}, a causal 1D CNN with dilated convolutions and residual blocks. TCNs are known to outperform recurrent models on sequence modeling tasks due to their parallelism and ability to model long-range dependencies efficiently. We adopt the generic configuration evaluated in~\cite{tcn_paper} with an added classification head and adjust the number of levels, dilation rates, and kernel sizes.

\section{Experimental Results} \label{sec:exp_results}

This section presents the experiments and results on our multi-modal dataset. We evaluate all models on three CSI configurations: 5 GHz @10 Hz (low-rate), 5 GHz @200 Hz (high-rate), and 60 GHz mmWave @10 Hz. 

\begin{table*}[hbt!]
\centering
% \captionsetup{justification=centering,margin=0.4cm}
\caption{Person identification accuracy comparison across frequencies without background subtraction. Abbreviations: M = Mixup, GS = Gaussian Smoothing, BiDi = Bidirectional, DR = Dropout Rate.}
\label{tab:freq_comp_without_bg}
\renewcommand{\arraystretch}{1.4}
\begin{tabular}{|l|l|c|c|c|c|}
\hline
\textbf{Model} & \textbf{Configuration} & \textbf{\#Params} & \textbf{5 GHz @10Hz} & \textbf{5 GHz @200Hz} & \textbf{60 GHz @10Hz} \\
\hline
\hline

CustomResNet1D & layers=[1,1,1,1], M & 1.9M & $\mathbf{0.89 \pm 0.017}$ & $0.957 \pm 0.012$ & $0.293 \pm 0.168$ \\

% CustomResNet1D & input\_channels=52, M + GS & 1.90M & $0.893 \pm 0.021$ & $0.953 \pm 0.029$ & - \\
\hline
CustomECAResNet1D & layers=[1,1,1,1], M & 1.9M & $\mathbf{0.897 \pm 0.055}$ & $0.947 \pm 0.006$ & $0.43 \pm 0.135$ \\

% CustomECAResNet1D & input\_channels=52, M + GS & 1.90M & $0.847 \pm 0.021$ & $0.92 \pm 0.03$ & - \\
\hline
TemporalConvNet & [64,128], kernel\_size=2, DR=0.5, M & 79K & $\mathbf{0.91 \pm 0.02}$ & $0.943 \pm 0.025$ & $0.883 \pm 0.038$ \\

% TemporalConvNet & [64,128,128], kernel\_size=2, DR=0.5, M & 145K & $0.903 \pm 0.055$ & - & - \\
\hline
LSTM\_HumanFi & hidden\_dim=128, layers=2, DR=0.2, BiDi, GS & 587K & $\mathbf{0.843 \pm 0.023}$ & $0.62 \pm 0.036$ & $0.6 \pm 0.062$ \\

% LSTM\_HumanFi & hidden\_dim=128, layers=1, DR=0.2, BiDi, GS & 192K & $0.853 \pm 0.021$ & - & - \\
\hline
CNN-BiLSTM (Temp. Attention) & hidden\_dim=64, layers=1, GS & 79K & $\mathbf{0.83 \pm 0.03}$ & $0.833 \pm 0.067$ & $0.827 \pm 0.04$ \\

% CNN-BiLSTM (Temp. Attention) & hidden\_dim=64, layers=2, M & 179K & $0.877 \pm 0.012$ & - & - \\
\hline
CNN-BiLSTM (Dual Attention) & lstm\_units=64, layers=2, GS & 188K & $\mathbf{0.81 \pm 0.035}$ & $0.677 \pm 0.042$ & $0.793 \pm 0.085$ \\

% CNN-BiLSTM (Dual Attention) & lstm\_units=64, layers=1 & 89K & $0.85 \pm 0.017$ & - & - \\
\hline
OptECAResNet1D-JARIL & layers=[1,1,1,1], GS & 3.64M & $\mathbf{0.857 \pm 0.029}$ & $0.957 \pm 0.012$ & $0.840 \pm 0.113$ \\

% OptECAResNet1D-JARIL & layers=[1,1,1,1] & 3.64M & $0.723 \pm 0.061$ & $0.957 \pm 0.032$ & $0.663 \pm 0.216$ \\
\hline
OptResNet1D-JARIL & layers=[1,1,1,1], GS & 7.07M & $0.83 \pm 0.036$ & $0.913 \pm 0.032$ & $\mathbf{0.867 \pm 0.071}$ \\

% OptResNet1D-JARIL & layers=[1,1,1,1] & 7.06M & $0.803 \pm 0.006$ & - & $0.46 \pm 0.43$ \\

\hline
\hline

CustomResNet1D & layers=[1,1,1,1], M + GS & 1.9M & $\mathbf{0.893 \pm 0.021}$ & $0.953 \pm 0.029$ & $0.3 \pm 0.155$ \\

% CustomResNet1D & input\_channels=52 & 1.90M & $0.857 \pm 0.045$ & $0.953 \pm 0.015$ & $0.833 \pm 0.110$ \\
\hline
CustomECAResNet1D & layers=[1,1,1,1], M + GS & 1.9M & $\mathbf{0.847 \pm 0.021}$ & $0.92 \pm 0.03$ & $0.383 \pm 0.071$ \\

% CustomECAResNet1D & input\_channels=52, GS & 1.90M & $0.867 \pm 0.006$ & $0.963 \pm 0.012$ & $0.387 \pm 0.202$ \\
\hline
TemporalConvNet & [64,128,128], kernel\_size=3, DR=0.2, M & 210K & $\mathbf{0.897 \pm 0.006}$ & $0.937 \pm 0.015$ & $0.857 \pm 0.012$ \\

% TemporalConvNet & [64,128], kernel\_size=3, DR=0.2, M + GS & 111K & $0.893 \pm 0.006$ & $0.937 \pm 0.015$ & - \\
\hline
LSTM\_HumanFi & hidden\_dim=128, layers=1, DR=0.5, BiDi & 191K & $\mathbf{0.84 \pm 0.03}$ & $0.733 \pm 0.055$ & $0.647 \pm 0.051$ \\

% LSTM\_HumanFi & hidden\_dim=128, layers=1, DR=0.2, M & 191K & $0.84 \pm 0.044$ & $0.693 \pm 0.015$ & - \\
\hline
CNN-BiLSTM (Temp. Attention) & hidden\_dim=128, layers=2, M & 611K & $\mathbf{0.83 \pm 0.035}$ & $0.917 \pm 0.04$ & $0.77 \pm 0.046$ \\

% CNN-BiLSTM (Temp. Attention) & hidden\_dim=64, layers=1, M + GS & 79K & $0.843 \pm 0.023$ & $0.93 \pm 0.026$ & - \\
\hline
CNN-BiLSTM (Dual Attention) & lstm\_units=64, layers=1, GS & 89K & $\mathbf{0.867 \pm 0.006}$ & $0.75 \pm 0.066$ & $0.813 \pm 0.060$ \\

% CNN-BiLSTM (Dual Attention) & lstm\_units=128, layers=1, M & 272K & $0.813 \pm 0.025$ & $0.753 \pm 0.067$ & - \\
\hline
OptECAResNet1D-JARIL & layers=[1,1,1,1] & 3.64M & $\mathbf{0.723 \pm 0.061}$ & $0.957 \pm 0.032$ & $0.663 \pm 0.216$ \\

% OptECAResNet1D-JARIL & layers=[1,1,1,1], M + GS & 3.64M & $0.823 \pm 0.061$ & $0.937 \pm 0.042$ & - \\
\hline
OptResNet1D-JARIL & layers=[1,1,1,1], M + GS & 7.07M & $\mathbf{0.76 \pm 0.035}$ & $0.917 \pm 0.025$ & $0.39 \pm 0.076$ \\
% OptResNet1D-JARIL & layers=[1,1,1,1], GS & 7.06M & - & $0.913 \pm 0.032$ & - \\

\hline
\hline

CustomResNet1D & layers=[1,1,1,1] & 1.9M & $\mathbf{0.857 \pm 0.045}$ & $0.953 \pm 0.015$ & $0.833 \pm 0.110$ \\

% CustomResNet1D & input\_channels=60, GS & 1.91M & $0.9 \pm 0.01$ & - & $0.390 \pm 0.301$ \\
\hline
CustomECAResNet1D & layers=[1,1,1,1], GS & 1.9M & $\mathbf{0.867 \pm 0.006}$ & $0.963 \pm 0.012$ & $0.387 \pm 0.202$ \\

% CustomECAResNet1D & input\_channels=60 & 1.91M & $0.847 \pm 0.051$ & - & $0.567 \pm 0.363$ \\
\hline
TemporalConvNet & [64,128], kernel\_size=3, DR=0.5, M & 113K & $\mathbf{0.92 \pm 0.035}$ & $0.95 \pm 0.017$ & $0.887 \pm 0.032$ \\

% TemporalConvNet & [64,128,128], kernel\_size=2, DR=0.2, M & 147K & $0.913 \pm 0.025$ & - & $0.813 \pm 0.049$ \\
\hline
LSTM\_HumanFi & hidden\_dim=64, layers=2, BiDi, DR=0.5, M & 166K & $\mathbf{0.84 \pm 0.017}$ & $0.647 \pm 0.045$ & $0.807 \pm 0.045$ \\

% LSTM\_HumanFi & hidden\_dim=128, layers=2, DR=0.2, M + GS & 232K & $0.823 \pm 0.050$ & - & $0.793 \pm 0.029$ \\
\hline
CNN-BiLSTM (Temp. Attention) & hidden\_dim=64, layers=2, M + GS & 180K & $\mathbf{0.81 \pm 0.05}$ & $0.903 \pm 0.015$ & $0.690 \pm 0.137$ \\

% CNN-BiLSTM (Temp. Attention) & hidden\_dim=64, layers=1, M & 81K & $0.85 \pm 0.053$ & - & $0.79 \pm 0.19$ \\
\hline
CNN-BiLSTM (Dual Attention) & lstm\_units=128, layers=2 & 669K & $\mathbf{0.833 \pm 0.031}$ & $0.677 \pm 0.11$ & $0.823 \pm 0.045$ \\

% CNN-BiLSTM (Dual Attention) & lstm\_units=128, layers=1, GS & 273K & $0.83 \pm 0.035$ & - & $0.837 \pm 0.038$ \\
% \hline
% OptECAResNet1D-JARIL & layers=[1,1,1,1] & 3.64M & - & - & $0.663 \pm 0.216$ \\
% OptECAResNet1D-JARIL & layers=[1,1,1,1], GS & 3.64M & - & - & $0.840 \pm 0.113$ \\
\hline
OptResNet1D-JARIL & layers=[1,1,1,1] & 7.07M & $\mathbf{0.803 \pm 0.006}$ & $0.93 \pm 0.036$ & $0.46 \pm 0.43$ \\
% OptResNet1D-JARIL & layers=[1,1,1,1], GS & 7.07M & - & - & $0.867 \pm 0.071$ \\
\hline

\end{tabular}
\end{table*}

\subsection{Without Background Subtraction}

Table~\ref{tab:freq_comp_without_bg} presents the top-performing configuration for each model class across three settings: low-rate 5 GHz, high-rate 5 GHz, and low-rate 60 GHz, all without background subtraction. The table is divided into three horizontal sections: models optimized for low-rate 5 GHz, high-rate 5 GHz, and 60 GHz, respectively. For each configuration, we report test accuracy across all three settings to illustrate how well models perform across them, even when originally optimized for another. To avoid redundancy, duplicate models are removed across sections, hence, not all sections contain eight entries. The same structure is used in Table~\ref{tab:freq_comp_with_bg}, which includes results with background subtraction. Bold entries in these two tabels indicate the top model between the two settings (5 GHz@10Hz and 60 GHz@10 Hz), enabling a fair cross-band comparison. Note that due to firmware constraints, we could not raise the 60 GHz CSI sampling frequency above 10 Hz without encountering instability issues.

We observe that increasing the sampling rate to 200 Hz significantly boosts performance for sub-6 GHz CSI across most models, confirming the value of higher sampling rate for capturing motion dynamics. However, the long input sequences (1000 time steps from 200 Hz over 5 seconds) of the high-rate 5 GHz appear to pose challenges for some LSTM-based models, potentially due to difficulties in capturing long-range dependencies or the risk of vanishing gradients over extended temporal contexts. On 60 GHz data, performance varies a lot across architectures. Although some models (e.g., \texttt{TemporalConvNet}) achieve relatively high accuracy, it is still lower than for the other frequencies. Other models perform inconsistently and exhibit high variance. Based on the 60 GHz results in Section~\ref{sec:freq_comp_with_bgsub}, we primarily attribute this performance gap to the absence of background subtraction, with a more detailed explanation provided in that section.

\begin{table*}[hbt!]
\centering
% \captionsetup{justification=centering,margin=0.4cm}
\caption{Person identification accuracy comparison across frequencies with background subtraction. Abbreviations: M = Mixup, GS = Gaussian Smoothing, BiDi = Bidirectional, DR = Dropout Rate.}
\label{tab:freq_comp_with_bg}
\renewcommand{\arraystretch}{1.4}
\begin{tabular}{|l|l|c|c|c|c|}
\hline
\textbf{Model} & \textbf{Configuration} & \textbf{\#Params} & \textbf{5 GHz @10Hz} & \textbf{5 GHz @200Hz} & \textbf{60 GHz @10Hz} \\
\hline
\hline

CustomResNet1D & layers=[1,1,1,1], M & 1.9M & $\mathbf{0.907 \pm 0.021}$ & $0.923 \pm 0.029$ & $0.85 \pm 0.017$\\

% CustomResNet1D & input\_channels=52, M + GS & 1.90M & $0.893 \pm 0.021$ & $0.953 \pm 0.029$ & - \\
\hline
CustomECAResNet1D & layers=[1,1,1,1], M & 1.9M & $\mathbf{0.897 \pm 0.012}$ & $0.943 \pm 0.006$ & $0.887 \pm 0.012$\\

% CustomECAResNet1D & input\_channels=52, M + GS & 1.90M & $0.847 \pm 0.021$ & $0.92 \pm 0.03$ & - \\
\hline
TemporalConvNet & [64,128], kernel\_size=2, DR=0.2, M & 79K & $0.91 \pm 0.017$ & $0.933 \pm 0.012$ & $\mathbf{0.963 \pm 0.006}$ \\

% TemporalConvNet & [64,128,128], kernel\_size=2, DR=0.5, M & 145K & $0.903 \pm 0.055$ & - & - \\
\hline
LSTM\_HumanFi & hidden\_dim=64, layers=1, DR=0.2, BiDi & 63K & $0.887 \pm 0.015$ & $0.647 \pm 0.029$ & $\mathbf{0.893 \pm 0.012}$ \\

% LSTM\_HumanFi & hidden\_dim=128, layers=1, DR=0.2, BiDi, GS & 192K & $0.853 \pm 0.021$ & - & - \\
\hline
CNN-BiLSTM (Temp. Attention) & hidden\_dim=64, layers=1, M + GS & 79K & $0.817 \pm 0.021$ & $0.87 \pm 0.03$ & $\mathbf{0.947 \pm 0.012}$ \\

% CNN-BiLSTM (Temp. Attention) & hidden\_dim=64, layers=2, M & 179K & $0.877 \pm 0.012$ & - & - \\
\hline
CNN-BiLSTM (Dual Attention) & lstm\_units=64, layers=1, GS & 89K & $0.843 \pm 0.015$ & $0.647 \pm 0.065$ & $\mathbf{0.9 \pm 0.017}$ \\

% CNN-BiLSTM (Dual Attention) & lstm\_units=64, layers=1 & 89K & $0.85 \pm 0.017$ & - & - \\
\hline
OptECAResNet1D-JARIL & layers=[1,1,1,1], GS & 3.64M & $0.86 \pm 0.05$ & $0.933 \pm 0.042$ & $\mathbf{0.903 \pm 0.031}$ \\

% OptECAResNet1D-JARIL & layers=[1,1,1,1] & 3.64M & $0.723 \pm 0.061$ & $0.957 \pm 0.032$ & $0.663 \pm 0.216$ \\
\hline
OptResNet1D-JARIL & layers=[1,1,1,1], GS & 7.06M & $0.85 \pm 0.035$ & $0.853 \pm 0.075$ & $\mathbf{0.907 \pm 0.049}$ \\

% OptResNet1D-JARIL & layers=[1,1,1,1] & 7.06M & $0.803 \pm 0.006$ & - & $0.46 \pm 0.43$ \\

\hline
\hline

CustomResNet1D & layers=[1,1,1,1], GS & 1.9M & $0.853 \pm 0.040$ & $0.893 \pm 0.074$ & $\mathbf{0.947 \pm 0.012}$ \\

% CustomResNet1D & input\_channels=52 & 1.90M & $0.857 \pm 0.045$ & $0.953 \pm 0.015$ & $0.833 \pm 0.110$ \\
\hline
CustomECAResNet1D & layers=[1,1,1,1], GS & 1.9M & $0.847 \pm 0.012$ & $0.943 \pm 0.006$ & $\mathbf{0.937 \pm 0.006}$ \\

% CustomECAResNet1D & input\_channels=52, GS & 1.90M & $0.867 \pm 0.006$ & $0.963 \pm 0.012$ & $0.387 \pm 0.202$ \\
\hline
TemporalConvNet & [64,128,256], kernel\_size=2, DR=0.2, M + GS & 312K & $0.86 \pm 0.017$ & $0.953 \pm 0.015$ & $\mathbf{0.923 \pm 0.032}$ \\

% TemporalConvNet & [64,128], kernel\_size=3, DR=0.2, M + GS & 111K & $0.893 \pm 0.006$ & $0.937 \pm 0.015$ & - \\
\hline
LSTM\_HumanFi & hidden\_dim=64, layers=1, DR=0.5, BiDi & 63K & $0.863 \pm 0.006$ & $0.657 \pm 0.068$ & $\mathbf{0.913 \pm 0.015}$ \\

% LSTM\_HumanFi & hidden\_dim=128, layers=1, DR=0.2, M & 191K & $0.84 \pm 0.044$ & $0.693 \pm 0.015$ & - \\
% \hline
% CNN-BiLSTM (Temp. Attention) & hidden\_dim=64, layers=1, M + GS & 79K & $0.817 \pm 0.021$ & $0.87 \pm 0.03$ & $0.947 \pm 0.012$ \\

% CNN-BiLSTM (Temp. Attention) & hidden\_dim=64, layers=1, M + GS & 79K & $0.843 \pm 0.023$ & $0.93 \pm 0.026$ & - \\
\hline
CNN-BiLSTM (Dual Attention) & lstm\_units=128, layers=1 & 273K & $0.847 \pm 0.021$ & $0.753 \pm 0.040$ & $\mathbf{0.91 \pm 0.03}$ \\

% CNN-BiLSTM (Dual Attention) & lstm\_units=128, layers=1, M & 272K & $0.813 \pm 0.025$ & $0.753 \pm 0.067$ & - \\
% \hline
% OptECAResNet1D-JARIL & layers=[1,1,1,1], GS & 3.64M & $0.86 \pm 0.05$ & $0.933 \pm 0.042$ & $0.903 \pm 0.031$ \\

% OptECAResNet1D-JARIL & layers=[1,1,1,1], M + GS & 3.64M & $0.823 \pm 0.061$ & $0.937 \pm 0.042$ & - \\
\hline
OptResNet1D-JARIL & layers=[1,1,1,1] & 7.06M & $0.747 \pm 0.041$ & $0.963 \pm 0.012$ & $\mathbf{0.853 \pm 0.015}$ \\
% OptResNet1D-JARIL & layers=[1,1,1,1], GS & 7.06M & - & $0.913 \pm 0.032$ & - \\

\hline
\hline

CustomResNet1D & layers=[1,1,1,1], M + GS & 1.9M & $0.9 \pm 0.0$ & $0.937 \pm 0.015$ & $\mathbf{0.92 \pm 0.046}$ \\

% CustomResNet1D & input\_channels=60, GS & 1.91M & $0.9 \pm 0.01$ & - & $0.390 \pm 0.301$ \\
\hline
CustomECAResNet1D & layers=[1,1,1,1], M + GS & 1.9M & $0.883 \pm 0.050$ & $0.93 \pm 0.026$ & $\mathbf{0.907 \pm 0.006}$ \\

% CustomECAResNet1D & input\_channels=60 & 1.91M & $0.847 \pm 0.051$ & - & $0.567 \pm 0.363$ \\
\hline
TemporalConvNet & [64,128,128], kernel\_size=2, DR=0.5, M & 147K & $0.93 \pm 0.0$ & $0.943 \pm 0.025$ & $\mathbf{0.947 \pm 0.012}$ \\

% TemporalConvNet & [64,128,128], kernel\_size=2, DR=0.2, M & 147K & $0.913 \pm 0.025$ & - & $0.813 \pm 0.049$ \\
\hline
LSTM\_HumanFi & hidden\_dim=128, layers=1, BiDi, DR=0.2, GS & 200K & $0.847 \pm 0.021$ & $0.623 \pm 0.032$ & $\mathbf{0.953 \pm 0.012}$ \\

% LSTM\_HumanFi & hidden\_dim=128, layers=2, DR=0.2, M + GS & 232K & $0.823 \pm 0.050$ & - & $0.793 \pm 0.029$ \\
\hline
CNN-BiLSTM (Temp. Attention) & hidden\_dim=128, layers=2, M + GS & 611K & $0.837 \pm 0.035$ & $0.87 \pm 0.044$ & $\mathbf{0.93 \pm 0.03}$ \\

% CNN-BiLSTM (Temp. Attention) & hidden\_dim=64, layers=1, M & 81K & $0.85 \pm 0.053$ & - & $0.79 \pm 0.19$ \\
\hline
CNN-BiLSTM (Dual Attention) & lstm\_units=128, layers=1, M + GS & 273K & $0.817 \pm 0.012$ & $0.727 \pm 0.057$ & $\mathbf{0.933 \pm 0.006}$ \\

% CNN-BiLSTM (Dual Attention) & lstm\_units=128, layers=1, GS & 273K & $0.83 \pm 0.035$ & - & $0.837 \pm 0.038$ \\
\hline
OptECAResNet1D-JARIL & layers=[1,1,1,1] & 3.64M & $0.767 \pm 0.025$ & $0.953 \pm 0.015$ & $\mathbf{0.897 \pm 0.038}$ \\
% OptECAResNet1D-JARIL & layers=[1,1,1,1], GS & 3.64M & - & - & $0.840 \pm 0.113$ \\
% \hline
% OptResNet1D-JARIL & layers=[1,1,1,1], GS & 7.06M & $0.85 \pm 0.035$ & $0.853 \pm 0.075$ & $0.907 \pm 0.049$ \\
% OptResNet1D-JARIL & layers=[1,1,1,1], GS & 7.07M & - & - & $0.867 \pm 0.071$ \\
\hline

\end{tabular}
\end{table*}

\begin{table*}[hbt!]
\centering
% \captionsetup{justification=centering,margin=0.4cm}
\caption{Comparison of 60 GHz vs. 5 GHz performance across all model configurations. Significant improvement ($\gg$) is defined as: $\mu_{60GHz} - \sigma_{60GHz} > \mu_{5GHz} + \sigma_{5GHz}$.}
\label{tab:freq_summary_stats_restructured}
\renewcommand{\arraystretch}{1.5}
\begin{tabular}{|l|c|c|c|}
\hline
\textbf{Metric} & \textbf{All Models (160)} & \textbf{Excl. LSTM-HumanFi (96)} & \textbf{Excl. All LSTM-based (64)} \\
\hline
Avg. Accuracy (60 GHz) & \textbf{0.9121} & \textbf{0.9148} & \textbf{0.9153} \\
Avg. Accuracy (5 GHz @10Hz) & 0.8432 & 0.8444 & 0.8584 \\
Avg. Accuracy (5 GHz @200Hz) & 0.7526 & 0.8536 & 0.9010 \\
\hline
60 GHz $>$ 5 GHz @10Hz (\# models) & \textbf{157} & \textbf{93} & \textbf{61} \\
60 GHz $>$ 5 GHz @200Hz (\# models) & \textbf{130} & \textbf{66} & \textbf{34} \\
\hline
60 GHz $\gg$ 5 GHz @10Hz (1 std) & \textbf{108} & \textbf{69} & \textbf{41} \\
60 GHz $\gg$ 5 GHz @200Hz (1 std) & \textbf{105} & 41 & 18 \\
\hline
\end{tabular}
\end{table*}

\subsection{With Background Subtraction} \label{sec:freq_comp_with_bgsub}

Table~\ref{tab:freq_comp_with_bg} reports the best person identification accuracy for all model classes across the three settings with background subtraction. We can see that, overall, background subtraction leads to improved or more stable accuracy for mmWave CSI (negligible for 5 GHz), for which most models exhibit high classification accuracy, often matching or surpassing their performance at high-rate 5 GHz. Notably, \texttt{TemporalConvNet} and \texttt{CNN-BiLSTM (Temp. Attention)} both achieve accuracies of 93–96\% on 60 GHz data, showing their ability to benefit from spatial detail. In contrast to the no-background-subtraction scenario, models like \texttt{CustomResNet1D} and \texttt{CustomECAResNet1D} also perform well at 60 GHz, showing that background removal plays an important role here.

The notable accuracy boost for 60 GHz with background subtraction likely stems from mmWave’s high spatial sensitivity. Its shorter wavelength captures fine environmental details, including static clutter that can obscure motion patterns; background subtraction reduces this noise, enhancing person-specific features. Conversely, background subtraction has little or negative effect on 5 GHz CSI, as its longer wavelength smooths out background clutter, making removal less beneficial. Notably, some LSTM-based models (e.g., \texttt{LSTM\_HumanFi}, \texttt{CNN-BiLSTM (Dual Attention)}) show marked 60 GHz performance gains over 5 GHz, especially at 200 Hz, reflecting better adaptation to mmWave signals.

\subsection{Aggregate Frequency Comparison Analysis}

To evaluate the overall performance difference between mmWave and 5 GHz Wi-Fi signals, we also did a comprehensive analysis across all trained model configurations, not only the best ones, using background-subtracted data. Table~\ref{tab:freq_summary_stats_restructured} summarizes the results, showing both average accuracies and counts of configurations where 60 GHz outperforms 5 GHz. Across all models, 60 GHz achieved the highest average accuracy (0.9121), outperforming low-rate 5 GHz in 157 out of 160 configurations, and the high-rate band in 130 configurations. To assess statistical significance, we used a stricter condition: 60 GHz was considered significantly better when its mean accuracy minus one standard deviation exceeded the 5 GHz accuracy plus its standard deviation. Under this criterion, 60 GHz significantly outperformed 5 GHz @10Hz in 108 configurations, and 5 GHz @200Hz in 105 configurations.
We further analyze results by excluding certain model categories. Without LSTM-HumanFi models (96 configurations), 60 GHz still achieves the highest average accuracy (0.9148) and is statistically superior in most cases. Excluding all LSTM models (64 configurations) maintains 60 GHz’s lead (0.9153), though 5 GHz at 200 Hz becomes more competitive, especially with CNNs. These results confirm that at comparable sampling rates, mmWave CSI outperforms traditional Wi-Fi sensing for person identification, with stronger statistical robustness. Note that we were unable to include the 200 Hz CSI rate for mmWave in our experiments due to firmware limitations of the COTS hardware. All configurations compared here and in Table~\ref{tab:freq_comp_with_bg} use background-subtracted data. Although this preprocessing step improves performance for 60 GHz CSI, it has little effect, or even slightly negative impact, on 5 GHz accuracy as we noticed. We can also implicitly look at 60 GHz with background subtraction against non-subtracted 5 GHz data using the two tables. Even in this comparison, 60 GHz still consistently outperforms low-rate 5 GHz. However, against high-rate 5 GHz, mmWave performance becomes more competitive but not really dominant, especially when considering non-LSTM-based architectures.

In summary, background subtraction substantially boosts the performance of mmWave CSI across models, allowing it to clearly outperform low-rate 5 GHz and approach the accuracy of high-rate 5 GHz. However, high-rate 5 GHz, particularly without background subtraction, can still surpass mmWave performance in certain model classes, especially CNN-based architectures. 

\subsubsection{Learning Curve Analysis} \label{sec:acc_vs_train}

\begin{figure}[t]
\centering
    \includegraphics[width=\columnwidth,trim = 0cm 3cm 12cm 0cm, clip]
    {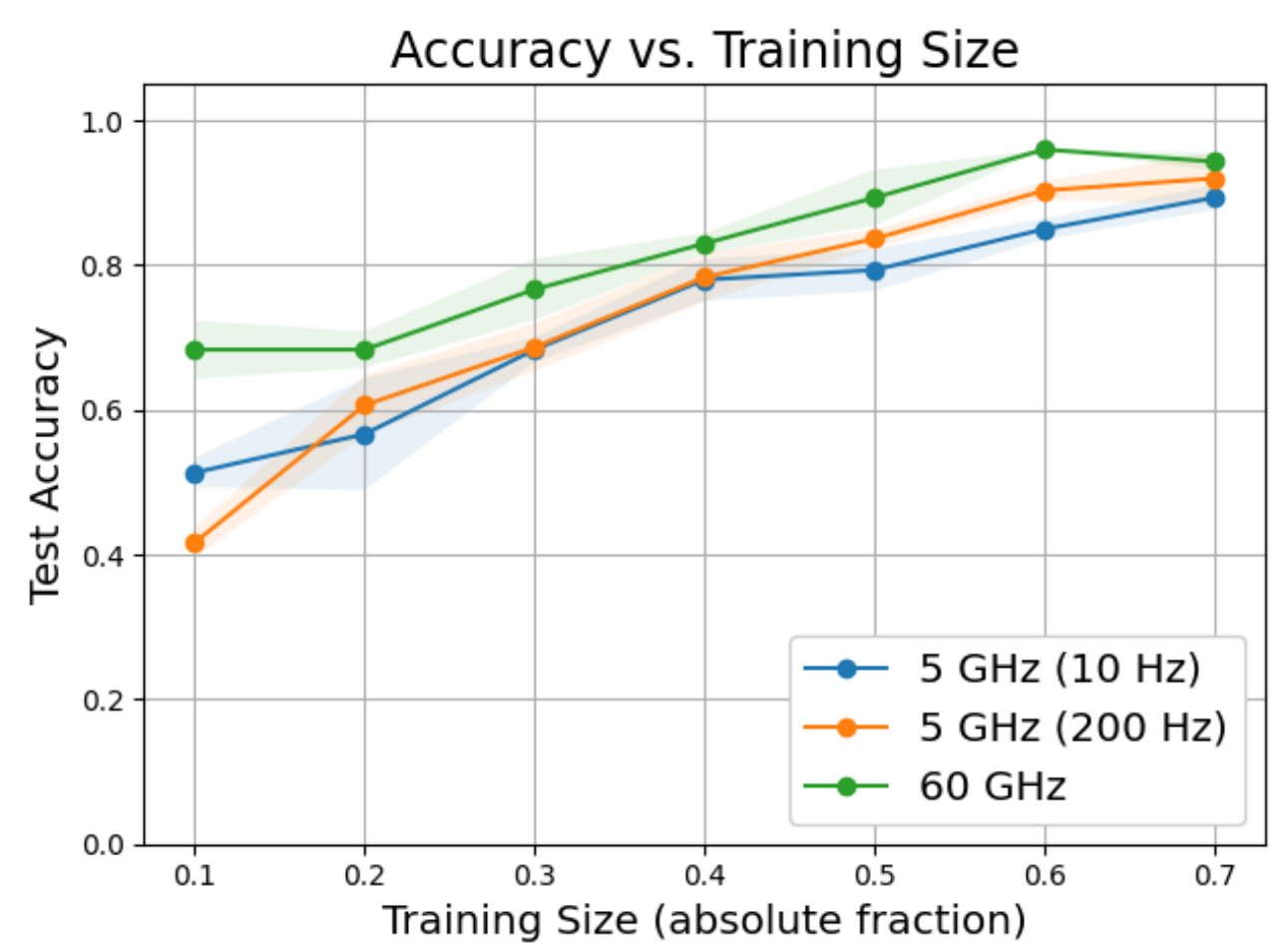}
    \caption{Learning curve for the \texttt{TemporalConvNet} model.}
    \label{fig:acc_vs_trainsize}
\end{figure}

To assess how frequency performance scales with training data size, we conduct a learning curve experiment. We select a representative model that demonstrates consistently strong performance across all frequency settings: the \texttt{TemporalConvNet} architecture, configured with three temporal convolutional blocks of widths (filters) \([64, 128, 128]\), a kernel size of 2, dropout rate of 0.5, and trained with Mixup data augmentation. This model is trained on progressively smaller portions of the training data (i.e., 70\%, 60\%, ..., down to 10\% of the full dataset), while performance is evaluated on a fixed test set (15\% of the total dataset). 
% The experiment is repeated for sample durations of 2, 3, and 5 seconds to also evaluate how temporal context influences identification accuracy. 
Fig.~\ref{fig:acc_vs_trainsize} presents the learning curves, with the shaded regions indicating the standard deviations. The 60 GHz data consistently outperforms both 5 GHz variants, showing better robustness to reduced training data. The performance gap increases significantly at smaller training fractions (below 30\%). Thus mmWave-based person identification shows better robustness under limited training data.

% \subsubsection{Effect of Number of Participants} \label{sec:varying_people_eval}

% We also evaluate how well the frequencies generalize as the number of target classes (i.e., participants) decreases in the total dataset. We use subsets of 5, 10, 15, and 20 participants, and evaluate performance at each level. Note that when we use a subset that contains less than 20 participants, they are picked randomly from the total pool of participants. Since we run this for multiple seeded runs, as we do everywhere, the random subset is also different for every seed. This is again repeated for multiple sample durations.

% Fig.~\ref{fig:acc_vs_people} shows how person identification accuracy scales with the number of participants. For 2-second and 3-second sample durations, 60 GHz shows comparable or superior performance compared to 5 GHz, especially at higher participant counts (10–20 people). For the 5-second sample case, 60 GHz performs worse overall when compared to 5 GHz, except when the participant count reaches 20. Regarding 5GHz variants, 10Hz only outperforms 200Hz at 5 participants, with 200Hz demonstrating better accuracy for larger groups.

% This suggests that 60 GHz CSI becomes more advantageous compared to 5GHz as the number of individuals grows, likely due to its ability to capture subtle spatial patterns that distinguish many similar gaits. However, for smaller group sizes, simpler signals with lower resolution (i.e., 5 GHz) may suffice.

\section{Conclusion and Future Work} \label{sec:conclusion}

In this study, we investigated Wi-Fi-based person identification using a novel multi-modal dataset comprising synchronized 5 GHz and 60 GHz CSI measurements from 20 participants in an indoor environment. Collected entirely with COTS hardware, the dataset enables direct comparison between frequency bands under identical conditions. Experimental results show that 60 GHz signals, when combined with background subtraction, yield the highest accuracy (91.2\% on average), highlighting the benefits of mmWave sensing in typical indoor settings and data-limited scenarios. However, high-rate 5 GHz CSI also performs competitively, suggesting that increased temporal resolution can partially offset the spatial limitations of sub-6 GHz systems. 

In future work, we intend to investigate fusion techniques to explore the combination of 5 GHz and 60 GHz CSI for enhanced identification performance. Additionally, we plan to examine higher mmWave sampling rates, which were constrained by device firmware in this work, with the expectation that increased temporal resolution could further improve performance.

\section{Acknowledgments}
Nabeel Bhat is funded by the Fund for Scientific Research Flanders (FWO) under grant number
 1SH5X24N. Part of this work is funded by the FWO WaveVR project
(Grant number: G034322N). 
 % Part of this work was supported by the Hexa-X-II project which has received funding from the Smart Networks and Services Joint Undertaking (SNS JU) under the European Union’s Horizon Europe research and innovation
% programme under Grant Agreement No 101095759

% \newpage
% \hfill\break
% \hfill\breaks
\bibliographystyle{ieeetr}
\bibliography{references}
\end{document}